# Fuzzy Rule Interpolation Toolbox for the GNU Open-Source OCTAVE

Maen Alzubi*, Mohammad Almseidin*, Mohd Aaqib Lone* and Szilveszter Kovacs*

* Department of Information Technology, University of Miskolc, H-3515 Miskolc, Hungary

* Email: alzubi@iit.uni-miskolc.hu, alsaudi@iit.uni-miskolc.hu, mohdaaqiblone@iit.uni-miskolc.hu, szkovacs@iit.uni-miskolc.hu

*Abstract*—In most fuzzy control applications (applying classical fuzzy reasoning), the reasoning method requires a complete fuzzy rule-base, i.e all the possible observations must be covered by the antecedents of the fuzzy rules, which is not always available. Fuzzy control systems based on the Fuzzy Rule Interpolation (FRI) concept play a major role in different platforms, in case if only a sparse fuzzy rule-base is available. This cases the fuzzy model contains only the most relevant rules, without covering all the antecedent universes. The first FRI toolbox being able to handle different FRI methods was developed by Johanyak et. al. in 2006 for the MATLAB environment. The goal of this paper is to introduce some details of the adaptation of the FRI toolbox to support the GNU/OCTAVE programming language. The OCTAVE Fuzzy Rule Interpolation (OCTFRI) Toolbox is an open-source toolbox for OCTAVE programming language, providing a large functionally compatible subset of the MATLAB FRI toolbox as well as many extensions. The OCTFRI Toolbox includes functions that enable the user to evaluate Fuzzy Inference Systems (FISs) from the command line and from OCTAVE scripts, read/write FISs and OBS to/from files, and produce a graphical visualisation of both the membership functions and the FIS outputs. Future work will focus on implementing advanced fuzzy inference techniques and GUI tools.

## I. INTRODUCTION

The most research papers related to Fuzzy Rule Interpolation (FRI) methodology are apparently created using the MATLAB environment. Unfortunately, there is a lack of FRI toolbox implementation in other programming languages. There are many implementations of the classical fuzzy reasoning methods for dense rule base systems, but the Matlab FRI toolbox [1] is the only FRI methodology collection for sparse rule bases. It implements a set of different methods, because up to now, there is no common FRI methodology for handling sparse fuzzy models. More techniques exist which follows each other's, but none of them is accepted as universal FRI solution. There is a need for a toolbox in which any of these methods can be applied and compared.

The first FRI Toolbox was implemented by Johanyak et al´ for the MATLAB environment in [1]. This special toolbox is a collection of MATLAB FRI functions, which can be run under MATLAB. It is easy to use and handy tool for demonstration and research purposes, but the integration into the real application is troublesome. Therefore, a new developer framework which includes more FRI methods has been designed and implemented.

This paper focuses on the use of the Fuzzy Rule Interpolation (FRI) to support GNU/OCTAVE program. The OCTAVE Fuzzy Rule Interpolation (OCTFRI) toolbox is an open-source toolbox for OCTAVE that provides a large MATLAB compatible subset of the functionality of the MATLAB FRI toolbox as well as many extensions. The OCTFRI toolbox includes functions that enable the user to evaluate Fuzzy Inference Systems (FISs) from the command line and from OCTAVE scripts, read/write FISs and OBS to/from files, and produce a graphical output of both the membership functions and the FIS outputs. Future work will focus on implementing advanced fuzzy inference techniques and GUI tools. The OCTAVE FRI toolbox is available for download on [2].

The rest of the paper is organized as follows: Section II provides the background of the FRI toolbox. Section III presents features of the OCTAVE FRI toolbox and its structure and functions. Section IV introduces a summary of some Interpolative System (FIS) types. Section V presents related works of the FRI MATLAB toolbox. Section VI introduces the fuzzy rule interpolation OCTAVE toolbox itself. Section VII gives practical examples for the evaluation of two different FIS and OBS data files. The conclusion and future work are presented in section VIII.

## II. BACKGROUND

The use of the FRI toolbox for implementing type-1 fuzzy inference systems is no diffuse. Currently, there is a deficiency of the FRI toolbox available for creating other types of fuzzy systems and other programming languages. As mentioned before that the initial FRI toolbox was implemented in the MATLAB environment, in which researchers have opened and accessed the FRI source code, however, the use of source code still demands in-depth knowledge (i.e. fuzzy rule interpolation methods), and thus does not perform the same purpose as toolbox in terms of obtaining the material simply available for users from various backgrounds.

The FRI Toolbox is a set of FRI functions, which can be run under MATLAB. It is easy to use and beneficial tool for demonstration and research purpose, but in the real application could be troublesome. It is based on the MATLAB, that is restricted in terms of its availability (MATLAB is ruled to license fees) and its user base. These constraints, limit the appropriation practicable of the current MATLAB FRI toolbox. For this purpose, a new developer framework based on

OCTAVE environment which includes FRI methods has been introduced.

The FRI toolbox presented in this paper is based on the openly available OCTAVE language. As regarded, OCTAVE gives the advantage of not only being openly available but importantly that it, as a language is accessible to users from a broad variety of backgrounds. Moreover, the benefit of high-level mathematical languages such as OCTAVE and MATLAB provides built-in primitives for representing and manipulating vectors and matrices, which can be used to directly represent and manipulate fuzzy sets. Meanwhile, these languages provide rich built-in graphical abilities for 2D and 3D plotting, which can be utilized to represent fuzzy sets.

### III. FRI Toolbox Features

The OCTAVE FRI toolbox has several associations with the MATLAB FRI toolbox. This is natural as both FRI toolboxes perform determined FRI concepts and methods. The similarity between both languages, nevertheless, has the distinct benefit of greatly facilitating the use of either toolbox for researchers. The resemblance based on the essential FRI concept between both OCTAVE and MATLAB FRI toolboxes will be illustrated in III-A, followed by a basic overview of the main structure of FIS and OBS data files in III-B and the main function of the FRI displays in III-C.

#### A. The MATLAB and OCTAVE Program Languages

MATLAB and its toolboxes, including fuzzy logic and Fuzzy Rule Interpolation (FRI) toolbox have become in fact standards in both university and production environments for engineering computation and design. For many potential MATLAB users, however, the cost is prohibitive and the license too restrictive. To meet the needs of these users, an opensource equivalent of MATLAB, called OCTAVE, was written and shared under the GNU Public License [3]. OCTAVE is one of the high-level programming languages which aimed to work on numerical computations. It was developed to be a simple and interactive environment in comparison with other softwares, such as FORTRAN. OCTAVE is enhanced by usercontributed packages managed by OCTAVE-Forge [4]. The packages provide a wide-ranging of functionality.

One of the essential features of OCTAVE is its compatibility with MATLAB since it uses the same syntax used in MATLAB and it can run MATLAB m-files [5]. Programmers can find some differences between OCTAVE and MATLAB since OCTAVE parser allows some syntax, that MATLAB does not, such as quotes, where single and double quotes are supported in OCTAVE, but only single quotes are used in MATLAB. Another kind of differences is that OCTAVE supports C-style auto-increment and assignment operators where MATLAB does not.

Comments in MATLAB are written after % while in OCTAVE you can use both % and #. In MATLAB the control blocks (while, if and for) as well as the functions delimiter all finish with "end", while in OCTAVE you can also use "endwhile", "endif", "endfor" and "endfunction" respectively.

In MATLAB the not equal to operator is "=" while in OCTAVE "!=" is also valid. MATLAB does not accept increment operators such as "++" and "–", while OCTAVE accepts them. Therefore, this free, open-source implementation of established techniques in a MATLAB-like environment provides a practical alternative for many students and researchers.

While there is no difference between OCTAVE and MATLAB, in terms of the scope of the implementation the current state of the OCTAVE FRI toolbox is similar to the MATLAB FRI toolbox in terms of fuzzy rule interpolation type-1, however, the OCTAVE FRI toolbox. For the majority of the type-1 functionality, one can establish a comparison between the functions implemented in the OCTAVE FRI toolbox and those in the MATLAB FRI toolbox as is expected because of the essential FRI concept.

#### B. The FIS and OBS Structure

Because the FIS Structure is central to the OCTAVE fuzzy logic toolbox, the toolbox's FIS structure is designed to correspond to the data files in the MATLAB fuzzy logic toolbox. The FIS and OBS structure as implemented by MATLAB FRI. The OCTAVE fuzzy logic toolbox uses the same structure for its FIS and OBS. Fig. 1 shows the main structure of the FIS and OBS, also shows the member of the structure that describes the membership functions "paramsy", which includes the membership values for the characteristic points of the fuzzy sets in case of piecewise linear membership functions.

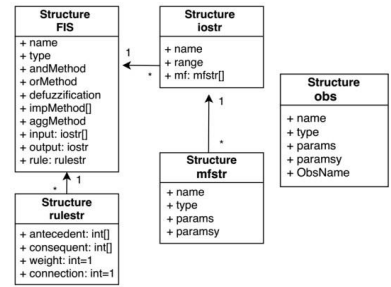

Fig. 1. FIS and OBS Data Structure [1]

The FIS and OBS data files format of the membership functions can be described as follows: (MF1='$A_{1;1}$':'trapmf', [3 8 12 17]![0 1 1 0]), (MF2='$A_{2;1}$':'trimf', [5 10 15]![0 1 0]), and (F3='$B_{1;1}$':'singlmf', [10]![1]), where, trimf, trapmf and singlmf refer to triangular, trapezoidal and singleton membership functions respectively, the $A_{1;1}$, $A_{2;1}$ and $B_{1;1}$ denote the names of the membership functions, the values [10 20 30], [4.5 5 5.5 6] and [0.46] denote to the characteristics points (params) of the membership in universe of discourse, a new parameter in fuzzy interpolation structure is called (paramsy) that was represented by ![0 1 0], ![0 1 1 0] and ![1], this parameter refer to the breakpoints of the characteristic points of the antecedents, consequences and observation fuzzy sets.

#### C. Implementation of ievalfis

The function "ievalfis" is the core of the FRI toolbox. "ievalfis" evaluates a FIS and OBS files of FRI methods, and for each set of inputs, ievalfis returns one or more crisp output values. The function "ievalfis" may be called as follows:

output=ievalfis(OBS, FIS, InterpolationType, Params);

The crisp input values are arranged in a matrix, called user input above, in which each row represents one set of FIS inputs. For an FIS that has M inputs, an input matrix of v sets of input values will have the form:

$$\begin{bmatrix} input_{11} & input_{12} & .. & input_{1M} \\ input_{21} & input_{22} & .. & input_{2M} \\ . & . & . & . \\ . & . & . & . \\ input_{v1} & input_{v2} & .. & input_{vM} \end{bmatrix}$$

The output is a matrix of crisp values in which each row represents the set of outputs for the corresponding row of user input. For an FIS that has M outputs, an output matrix corresponding to the preceding input matrix will have the form:

$$\begin{bmatrix} output_{11} & output_{12} & .. & output_{1M} \\ output_{21} & output_{22} & .. & output_{2M} \\ . & . & . & . \\ . & . & . & . \\ output_{v1} & output_{v2} & .. & output_{vM} \end{bmatrix}$$

The optional argument num-points specifies the number of points over which to evaluate the fuzzy values. If not specified, the default for some methods value of 101 or 501 in other methods is used. Furthermore, to returning the crisp output values, ievalfis returns its results based on the FRI method used. These results may be plotted or further analyzed. The format of the resulting output depends on the Fuzzy Interpolative System (FIS) type used.

## IV. FUZZY INTERPOLATIVE SYSTEM TYPES

This section presents a brief overview of different methods and approaches that are used for fuzzy rule interpolation concept. It also provides some relevant works related to the application of using the FRI MATLAB toolbox.

The first FRI method (KH FRI) was proposed by Koczy and´ Hirota [6]. It was initially referred as "linear interpolation", the KH FRI is based on linear interpolation of $\alpha$-cuts of the fuzzy sets, it applies linear interpolation for getting the conclusion for both end points of each $\alpha$-cuts of the observation, rule antecedents and consequent fuzzy sets. The conclusion could be calculated using the fundamental equation of the KH FRI (1), which is based on the lower and upper fuzzy distances between fuzzy sets [7].

$$dist(A*,A_1) : dist(A*,A_2) = dist(B*,B_1) : dist(B*,B_2) \quad (1)$$

where d refers to the Euclidean distance that could be used between the fuzzy sets $(A_2, A_2)$ and $(B_1, B_2)$.

The upper and lower endpoints could be used to calculate the distance between the conclusion and the consequent which must be comparable to the upper and lower fuzzy distances between observation and antecedents. It can be calculated by (2):

$$B^*_{\alpha C} = \frac{\sum_{i=1}^{2} \frac{B_{\alpha iC}}{d_C(A^*_{\alpha C}, A_{\alpha iC})}}{\sum_{k=1}^{2} \frac{1}{d_C(A^*_{\alpha C}, A_{\alpha kC})}} \quad (2)$$

A modification of the KH FRI is called (VKK FRI), it was proposed by Vass et al. [8]. This method is also following $\alpha$-cut technique, where the conclusion is computed based on the distance of the center points and the widths of the $\alpha$-cuts, instead of their lower and upper endpoints of the KH method as shown in Fig.2.

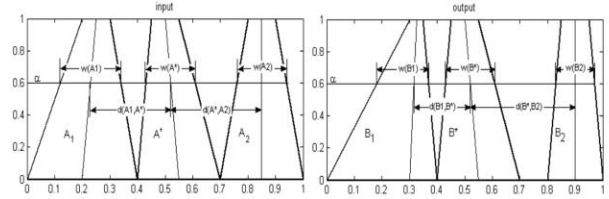

Fig. 2. The Distance of the Center Point and Width in the VKK Method [8]

The stabilized KH (KHstab FRI) was proposed by Tikk et al. [9] to handle and exclude the possible abnormal conclusion. For calculating the fuzzy conclusion, this method applies the reciprocal distance as weights between all the rule antecedents and the observation for all the $\alpha$-cuts. The universal approximation property holds if the distance function is raised to the power of the dimension of the input as shown by (3) and (4).

$$Min(B^*_\alpha) = \frac{\sum_{i=1}^{n} \frac{inf(B_{i\alpha})}{d_L^N(A^*_\alpha, A_{i\alpha})}}{\sum_{i=1}^{n} \frac{1}{d_L^N(A^*_\alpha, A_{i\alpha})}} \quad (3)$$

$$Max(B^*_\alpha) = \frac{\sum_{i=1}^{n} \frac{sup(B_{i\alpha})}{d_U^N(A^*_\alpha, A_{i\alpha})}}{\sum_{i=1}^{n} \frac{1}{d_U^N(A^*_\alpha, A_{i\alpha})}} \quad (4)$$

Another approach of FRI methods is called the Modified $\alpha$Cut based Interpolation (MACI) method [10]. The main idea of this method is based on the vectors description of the fuzzy sets for eliminating the abnormality problem in the conclusion. The fuzzy set in this method could be described by two vectors space, it can represent the left and the right flank of the $\alpha$cut levels where the abnormal consequent set is excluded. The conclusion B* in this method could be determined by (5).

$$RB* = (1 - \lambda core)RB1 + \lambda coreRB2 \quad (5)$$

Where,

$$\lambda core = \frac{\sqrt{\sum_{i=1}^{k}(RA^*_i - RA_{i1})^2}}{\sqrt{\sum_{i=1}^{k}(RA_{i2} - RA_{i1})^2}}$$

The Conservation Relative Fuzziness (CRF) method was introduced in [11] by Koczy, Hirota, and Gedeon. This method´

aims to obtain the conclusion based on determining the core of conclusion C* based on computing the ratio distances between fuzzy set of the antecedents ($d_1(A_1,A_2)$) and consequences ($d_1(B_1,B_2)$) with the core length of c* of the observation as shown in Fig.3. Therefore, the CRF technique determines the fuzziness of the (left $A_1$, right A*) must be the same fuzziness of the (left $B_1$, right B*), and similarity the fuzziness between (left A*, right $A_2$) and (left B*, right $B_2$).

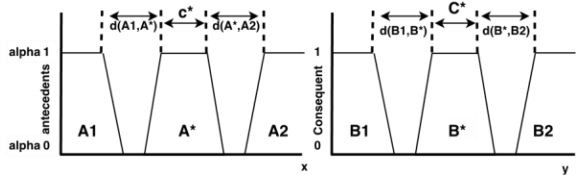

Fig. 3. The Ratio Distances Between Fuzzy Sets [11]

The IMUL method proposed by Wong, Gedeon and Tikk [12], IMUL was introduced to avoid the abnormal conclusion and improve the multidimensional α-cut (levels) based on the fuzzy interpolation. This method was introduced to combine between the features of MACI and CRF methods. The IMUL method applied the vector description, it can describe the characteristics points of the fuzzy sets through advantageous the transformation feature of MACI method, and representing the fuzziness of the input and output by CRF method. The conclusion in the IMUL method is based on calculating the left/right core (LC/RC) and flank (LF/RF) as using the Equations (6) and (7):

$$LCB* = (1 - \lambda_{core}).LCB_1 + \lambda_{core}.LCB_2 +$$
$$(\lambda_{core} - \lambda_{left/right}).(LCB_2 + LCB_1) \quad (6)$$

$$LFB^* = LCB^* - r_k \left(1 + \left|\frac{S'}{U'} - \frac{S}{U}\right|\right) \quad (7)$$

The General Method (GM), it was published by Baranyi et al. [13]. The GM method will adopt the characterization of the position fuzzy sets to determine the reference points (core), thus, the distance between the observation and antecedents fuzzy sets can be calculated based on the reference points as shown by Equation (8). The conclusion in this method could be determined by two algorithms, the first one is based on the fuzzy relation, which is to generate a new interpolated rule $R_i$ : $A_i \rightarrow B_i$ that positioned between rules $R_1$ and $R_2$, the position of the new rule is the same position of the observation, thus, each fuzzy set of the antecedents is used to produce the new rule which must be identical with the reference point of the observation fuzzy set in the corresponding dimension.

$$d(A_1,A_2) = |RP(A_2) - RP(A_1)| \quad (8)$$

The second one is based on the semantics of the relations, which is the new rule could be specified as a part of the extended rule of the approximate conclusion, as a conclusion of the inference method is defined by determining this rule. In many instances, there is no identical similarity between the rule and the observation part, for this purpose, many techniques are used to handle the mismatch by either the Transformation of the Fuzzy Relation (TFR) technique or by Fixed Point Law (FPL).

The ScaleMove method was presented by Huang and Shen [14], it follows the interpolation concept to handle the sparse fuzzy rule bases. The ScaleMove method provides the abilities to work with different fuzzy membership functions types. This method is based on the Centre Of Gravity (COG) of the membership functions as shown in Fig.4. It creates a new central rule via two adjacent rules that are surrounding the observation.

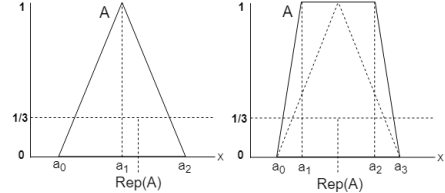

Fig. 4. Representative Value of a Triangular and Trapezoid Fuzzy Sets [14].

This ScaleMove method takes two steps to get the conclusion, the first step is to produce a new central rule ($A^0 \rightarrow B^0$) is produced within the existing surrounding rules between observation (A* : $A_1 \rightarrow B_1, A_2 \rightarrow B_2$) through to apply the Equation (9):

$$\lambda_{REP} = \frac{d(REP(A_1), REP(A^*))}{d(REP(A_1), REP(A_2))} \quad (9)$$

The second step is to calculate the $A^0$ similarity degree between fuzzy sets ($A^0$ and A*) that is transforming $B^0$ to B* with the desired degree of similarity by the scale and move transformations.

## V. FUZZY RULE INTERPOLATION MATLAB TOOLBOX AND RELATED WORKS

In [15], the fundamental concepts of a Fuzzy Rule Interpolation-based (FRI) Reinforcement Learning (RL) method called FRIQ-learning were discussed with benchmarks. The interpolation within the knowledge-base allows the removal of less important, unnecessary information, while still keeping the system functional. A Fuzzy Rule Interpolation based (FRI) RL method called FRIQ-learning is a method which possesses this feature. FRIQ-learning is also suitable for knowledge extraction. The FIVE FRI method was used By handling the antecedent and consequent fuzzy partitions of the fuzzy rule-base as scaling functions (weighting factors), which turns the fuzzy interpolation to scale crisp interpolation. The implementation of the FIVE FRI is also appearing in the FRI toolbox [2].

The authors in [16] give a brief introduction to the FRI methods and description of the refreshed and extended version of the original fuzzy rule interpolation MATLAB toolbox.
The methods used in the FRI toolbox (KH, KH Stabilized, MACI, IMUL, CRF, VKK, GM, FRIPOC, LESFRI, and

SCALEMOVE) were tested to compare them according to the abnormality and linearity criteria, based on different numerical benchmark examples.

The authors in [17] introduced the benefits of the FRI in the Intrusion Detection Systems (IDS) application area, for the design and implementation of the detection mechanism for Distributed Denial of Service (DDOS) attacks. The performance of the FRI-IDS application was compared to other common classification algorithms (support vector machine, neural network, random forest and decision tree) used for detecting DDOS attacks on the same open-source test-bed environment. According to the results, the overall detection rate of the FRI-IDS is in pair with other methods. Consequently, the FRI inference system could be a suitable approach to be implemented as a detection mechanism for IDS; as it effectively decreases the false positive rate value.

In [18], authors introduced a detection approach for defining abnormality by using the Fuzzy Rule Interpolation (FRI) methods with Simple Network Management Protocol (SNMP) Management Information Base (MIB) parameters. The implemented experiments were performed using Matlab and the Fuzzy Rule Interpolation Toolbox (FRIT) [2]. The FIVE method was chosen as the inference reasoning of the proposed detection approach.

Authors in [19], analyzed equations and notations related to piece-wise linearity property, which is aimed to highlight the problematic properties of the KH FRI method to prove its efficiency with piece-wise linearity condition. The study presented in the benchmark examples can be served as a baseline for testing other FRI methods against situations in which the linearity condition for KH FRI is not fulfilled. In the study, the FRI Toolbox was used to test the benchmark examples for different FRI methods.

## VI. FUZZY RULE INTERPOLATION OCTAVE (OCTFRI) TOOLBOX

### A. General description

The OCTAVE FRI toolbox is aimed to commonly giving at least the same set of features as it exists in the MATLAB FRI toolbox. Being an open-source toolbox, it is hoped that in the overall feature richness and functionality of the toolbox will be increased in time through the open access contribution.

The OCTFRI toolbox is the initial version of FRI techniques based on OCTAVE language, which includes FRI functions to evaluate FISs and OBSs files from the command line and OCTAVE scripts, which is done by read FISs and OBS files and produce a graphical output of both the membership functions and the FIS output. The main goal of the OCTFRI toolbox is to unify different fuzzy interpolation methods. The current version supports twelve FRI methods (KH, KHstabilized, MACI, IMUL, CRF, FIVE, VKK, GM, FRIPOC, LESFRI, VEIN, and ScaleMove (note that ScaleMove style inference is currently supported only in the OCTFRI toolbox)). Nevertheless, the number of included techniques is still growing. The toolbox is available for download under GNU general public license from website [2]. As part of describing the FRI inference systems functionality, a wide variety of forms of membership functions are supported similar to those currently provided by MATLAB such as singleton (singlmf), triangle (trimf), trapezoid (trapmf), polygon (polymf). Additionally, currently, the common forms of conjunction, disjunction, implication and aggregation operators are supported (including, minimum, maximum, product and probabilistic OR). The number of rules is not restricted. There are some restrictions to use FRI toolbox, only convex and normal fuzzy sets are allowed (see [20]). In case of GM the allowed set types are: singleton, triangle, trapezoid. Fig.5 describes the general structure of OCTFRI toolbox that could be used to run the OCTFRI toolbox and to evaluate the current FRI methods.

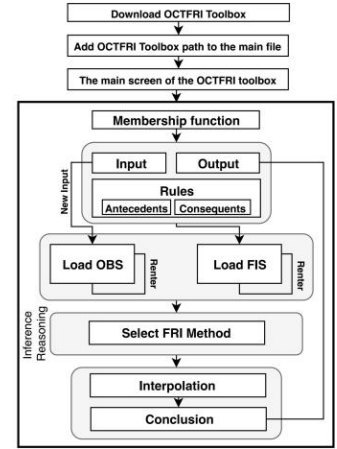

Fig. 5. The General Structure of OCTAVE FRI Toolbox

### B. Parameters of the FRI methods by OCTFRI toolbox

The FRI method parameters could be defined by the main file of the OCTFRI toolbox "OCTFRI.m". Regarding the KH, VKK and KHstabilized methods which use two types of parameters, the first parameter is "breakpoints" (0 or 1) is considered as a default parameter and denoted the $\alpha$ levels defined by the breakpoints [0 1] of the fuzzy sets, the second one is "userdefined", which the user specifies the number of $\alpha$ levels that will be distributed uniformly in the interval [0,1], for example:

    params.InterpolationType='KH';
    params.AlphaLevels.Type='breakpoints'; or
    params.AlphaLevels.Type='userdefined';

In case MACI and IMUL methods that used RPtype (corecentre) parameter, which refer to the type of the reference point of the fuzzy set, e.g. params.InterpolationType='MACI'; params.RPtype='corecentre'. For FRIPOC and LESFRI methods share the same parameters, where the FEAT-p technique can take all fuzzy sets that belong to the partition with different weight values. The type of the weighting factor and its parameters can also be set by the user, e.g.

    params.InterpolationType='FRIPOC';
    params.NumOfPoints=501; params.RPtype='corecentre';
    params.NumOfpCuts=61;

```
params.SetInterpolationWeight.p=2;
params.ConsequentPositionWeight.p=2;
```

Most of FRI methods calculate multidimensional distances in the Minkowski sense. The parameter w of the formula can also be set by the user where default value is 2. The default values of the FRI methods parameters are sufficient to show the desired conclusion.

*C. Usage of the OCTFRI toolbox and evolution an FIS and OBS*

The package of OCTFRI toolbox can be used from a graphical interface or from the command line. The current version of the OCTFRI is simple and easy to use, which contains all buttons of FRI methods, loading data files and interpolation testing as shown in Fig.6.

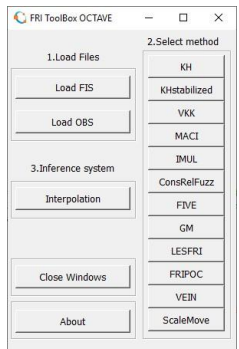

Fig. 6. The Main Screen of the OCTFRI Toolbox

OCTFRI toolbox can be begun by typing in the command line with the command "OCTFRI.m". First, the location of FIS and OBS data should be provided (see Fig.7), which can be done through the standard file open dialogue box.

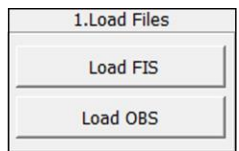

Fig. 7. Location of FIS and OBS Data Files

VII. EVALUATION A FIS AND OBS DATA FILES

Twelve separate functions of the FRI methods as presented in Fig.6 (eight of which were highlighted in this paper KH FRI, KHstab FRI, VKK FRI, ScaleMove FRI, MACI FRI, IMUL FRI, CRF FRI, and GM FRI) and six example FIS and OBS files are included in the toolbox. A FIS and an OBS files needs to be selected to be evaluated by the FRI method. The inference process starts by loading the FIS and OBS data and select one of the FRI methods, then the conclusion $B^*$ could be shown by pressing on the interpolation button (see Fig.6). The input and output universes will be shown in two separate windows, each including the same number of diagrams as the dimensions of the input and output respectively.

*A. Example 1: Evaluation of the KH, KHstab, VKK and the ScaleMove FRI Method*

Example 1 applies the FIS1 and OBS1 data files stored in an OCTFRI folder, which is including one input, one output dimensions, and only two fuzzy rules as shown in Fig.8. The membership functions of the input and output universes are triangular fuzzy sets. Example 1 is tested on the KH FRI [6], the KH Stabilized FRI [9], the VKK FRI [8], and the ScaleMove FRI [14]) methods as described in Figs. 9 and 10.

```
[System] %FIS FILE%              [Output1]
Name='FIS1'                      Name='output1'
Type='sparse'                    Range=[0 50]
Version=2.0                      NumMFs=2
NumInputs=1                      MF1='B1':'trimf',[5 10 15]![0 1 0]
NumOutputs=1                     MF2='B2':'trimf',[37 42 47]![0 1 0]
NumRules=2
AndMethod=''                     [Rules]
OrMethod=''                      1, 1 (1) : 1
ImpMethod=''                     2, 2 (1) : 1
AggMethod=''
DefuzzMethod='COG'               ***********************

[Input1]                         %OBS FILE%
Name='input1'                    NumInputs=1
Range=[0 50]                     ObsName='OBS1'
NumMFs=2                         [Observation]
MF1='A1':'trimf',[5 10 15]![0 1 0]  OBS1='A*_1':'trimf',[17 27 37]![0 1 0]
MF2='A2':'trimf',[37 42 47]![0 1 0]
```

Fig. 8. A FIS1 and OBS1 Stored in a File.

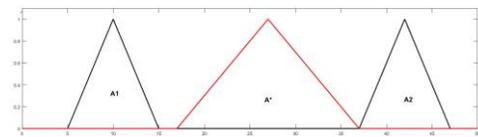

Fig. 9. Antecedent Partitions and Observations for the First Example.

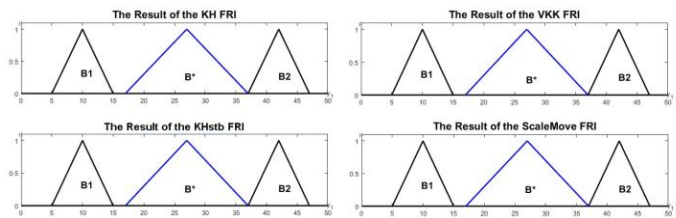

Fig. 10. The Consequent Partitions and Approximate Conclusions of the FRI Methods For the First Example.

*B. Example 2: Evaluation the MACI, CRF, IMUL and GM FRI methods*

Example 2 applies the FIS2 and OBS2 data files stored in an OCTFRI folder, which is including two input, one output dimensions, and only four fuzzy rules as shown in Fig.11. The membership functions of the input and output universes are triangular fuzzy sets. Example 2 is tested on the MACI FRI [10], the CRF FRI [21], the IMUL FRI [12], and the GM FRI [13]) methods as described in Figs. 12 and 13.

```
[System]  %FIS FILE%            [Output1]
Name='FIS2'                      Name='output1'
Type='sparse'                    Range=[0 80]
Version=2.0                      NumMFs=4
NumInputs=2                      MF1='B1':'trapmf',[3 8 12 17]![0 1 1 0]
NumOutputs=1                     MF2='B2':'trapmf',[23 28 32 37]![0 1 1 0]
NumRules=4                       MF3='B3':'trapmf',[43 48 53 58]![0 1 1 0]
AndMethod=''                     MF4='B4':'trapmf',[63 68 72 77]![0 1 1 0]
OrMethod=''
ImpMethod=''                     [Rules]
AggMethod=''                     1 1, 1 (1) : 1
DefuzzMethod='COG'               2 2, 2 (1) : 1
                                 3 3, 3 (1) : 1
[Input1]                         4 4, 4 (1) : 1
Name='input1'
Range=[0 80]
NumMFs=4                         ************************
MF1='mf1':'trapmf',[3 8 12 17]![0 1 1 0]   %OBS FILE%
MF2='mf2':'trapmf',[23 28 32 37]![0 1 1 0] NumInputs=2
MF3='mf3':'trapmf',[43 48 53 58]![0 1 1 0] ObsName='OBS2'
MF4='mf4':'trapmf',[63 68 72 77]![0 1 1 0] [Observation]
                                 OBS1='A*_1':'trapmf',[18 20 21 23]![0 1 1 0]
[Input2]                         OBS2='A*_2':'trapmf',[37 39 40 42]![0 1 1 0]
Name='input2'
Range=[0 80]
NumMFs=4
MF1='A1':'trapmf',[3 8 12 17]![0 1 1 0]
MF2='A2':'trapmf',[23 28 32 37]![0 1 1 0]
MF3='A3':'trapmf',[43 48 53 58]![0 1 1 0]
MF4='A4':'trapmf',[63 68 72 77]![0 1 1 0]
```

Fig. 11. A FIS2 and OBS2 Stored in a File.

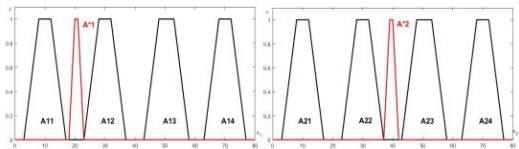

Fig. 12. Antecedent Partitions and Observations For the Second Example.

## VIII. CONCLUSIONS

This paper presented the initial version of the FRI toolbox ported to the OCTAVE language, which is an open-source (free license) software available under the (GPL). The OCTFRI toolbox is compatible with MATLAB through packages and syntax, and it can also run m files and make modifications without restrictions. The OCTFRI toolbox is a freely available public tool, which is designed to be a test-base for FRI methods comparison and also offers a simple way for solving real-world problems. The current OCTFRI toolbox implements twelve FRI methods (KH, KH Stabilized, VKK, MACI, IMUL, CRF, FIVE, GM, LESFRI, FRIPOC, VESI and ScaleMove). The SclaeMove method was added as a new inference method to the OCTAVE FRI. For demonstrating the usage of the OCTFRI, the results of two FIS and OBS data files were introduced in this paper to evaluate the conclusion of the KH, KH Stabilized, VKK, ScaleMove, MACI, CRF, IMUL and GM FRI methods by the OCTFRI toolbox.

### ACKNOWLEDGMENT

The described study was carried out as part of the EFOP-3.6.1-16-00011 Younger and Renewing University - Innovative Knowledge City - institutional development of the University of Miskolc aiming at intelligent specialization project implemented in the framework of the Szechenyi 2020 program. The realization of this project is supported by the European Union, co-financed by the European Social Fund.

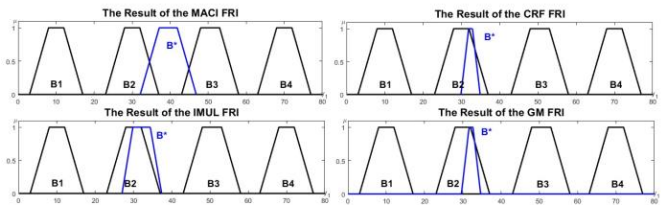

Fig. 13. The Consequent Partitions and Approximate Conclusions of the FRI Methods For the Second Example.